\documentclass{article}

     \usepackage[preprint,nonatbib]{Style_Ref/neurips_2020}

\usepackage[utf8]{inputenc} 
\usepackage[T1]{fontenc}    
\usepackage{hyperref}       
\usepackage{url}            
\usepackage{booktabs}       
\usepackage{amsfonts}       
\usepackage{nicefrac}       
\usepackage{microtype}      

\usepackage{float}
\usepackage[linesnumbered,ruled]{algorithm2e}
\usepackage{graphicx}
\usepackage{comment}
\usepackage{xcolor}
\usepackage{makecell,interfaces-makecell}
\usepackage{arydshln}
\usepackage{subcaption}
\usepackage{wrapfig}
\usepackage{amsfonts} 
\DeclareMathSymbol{\shortminus}{\mathbin}{AMSa}{"39}
\usepackage{amsmath}

\usepackage{multirow}

\title{Rethinking Class-Discrimination Based CNN Channel Pruning}

\author{%
  Yuchen~Liu, David~Wentzlaff, S.Y.~Kung
    \\
  Department of Electrical Engineering\\
  Princeton University\\
  \texttt{\{yl16, wentzlaf, kung\}@princeton.edu} \\
}

\begin{document}
\maketitle
\newcommand\metric{G-SD}
\newcommand\analysis{FLOP-normalized sensitivity analysis}

\begin{abstract}
Channel pruning has received ever-increasing focus on network compression. 
In particular, class-discrimination based channel pruning has made major headway, 
as it fits seamlessly with the classification objective of CNNs and provides good explainability. 
Prior works singly propose and evaluate their discriminant functions, 
while further study on the effectiveness of the adopted metrics is absent.
To this end, we initiate the first study on the effectiveness of a broad range of discriminant functions on channel pruning.
Conventional single-variate binary-class statistics like Student's T-Test are also included in our study via an intuitive generalization.
The winning metric of our study has a greater ability to select informative channels over other state-of-the-art methods,
which is substantiated by our qualitative and quantitative analysis.
Moreover, we develop a \analysis\ scheme to automate the structural pruning procedure. 
On CIFAR-10, CIFAR-100, and ILSVRC-2012 datasets, our pruned models achieve higher accuracy with less inference cost compared to state-of-the-art results. 
For example, on ILSVRC-2012, 
our 44.3\% FLOPs-pruned ResNet-50 has only a 0.3\% top-1 accuracy drop, which significantly outperforms the state of the art.


\end{abstract}
\section{Introduction}
\label{sec:intro}

Convolutional neural networks (CNNs) have become a mainstream machine learning model for various computer vision tasks, such as 
image classification~\cite{he2016deep, simonyan2014very}, 
image super resolution~\cite{ledig2017photo, wang2018esrgan}, and 
semantic segmentation~\cite{long2015fully, ronneberger2015u}.
To gain better prediction performance, a popular approach is to grow deeper and wider models.
However, such CNNs require larger storage space and higher computational cost, 
making them unsuitable for edge devices like phones and embedded sensors. 

Many methods have been proposed to reduce the storage space and inference latency of CNNs. 
For example: weight quantization~\cite{chen2015compressing,courbariaux2016binarized}, 
tensor low-rank factorization~\cite{jaderberg2014speeding,lebedev2014speeding},  
weight-level pruning~\cite{han2015learning, han2015deep, hassibi1993second}, 
and channel-level pruning~\cite{ zhuang2018discrimination, he2019filter, yu2018nisp, kung2019methodical}.  
Among them all, channel-level pruning is the preferable method 
as it produces smaller dense models 
which can easily leverage high-efficiency Basic Linear Algebra Subprograms (BLAS) libraries~\cite{li2016pruning}.

\begin{figure}[H]
    \centering
    \includegraphics[width=0.9\textwidth]{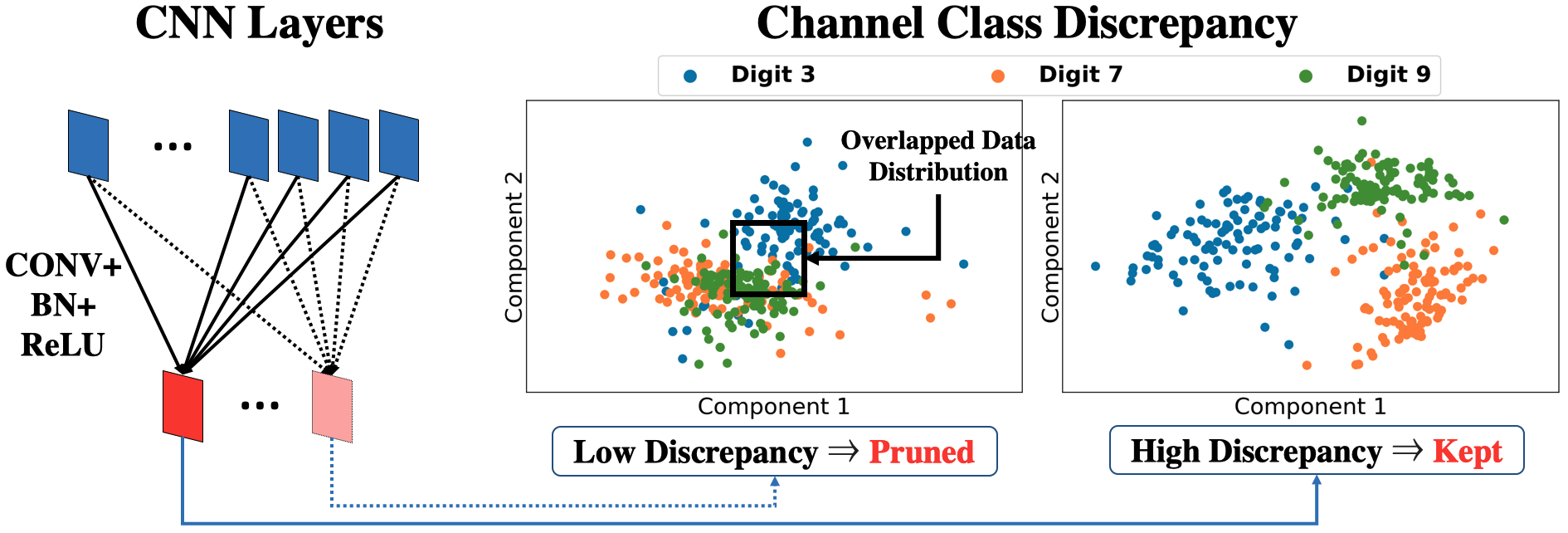}
    \caption{ An illustration of discrimination-based channel pruning. 
    Feature maps of two channels are picked from the second layer of LeNet-5~\cite{lecun1998gradient} fed with MNIST~\cite{lecun1998mnist} input.
    The pruning decision is visually explainable, as shown in feature maps' 2D projection.
    The channel with a low discrimination score (in dashed) is discarded as it has little class discrepancy. 
    In contrast, the channel with a higher score displays clearer class separability and is kept during the pruning process. (Best viewed in color)}
    \label{fig:intro_idea}
\end{figure}

Channel pruning has recently been fostered by the notion of evaluating the class-discriminative information 
with state-of-the-art works~\cite{zhuang2018discrimination,kung2019methodical}.
Such methodology has two unique advantages:
(1) It aims to find channels that have great linear separability with respect to the labels,
which fits seamlessly with the goal of classification CNNs, to transform original distribution into a linearly separable space~\cite{haykin1994neural}. 
(2) It makes the pruning process more explainable, as it directly assesses what the channels have learned and quantitates the information of the channels in rigorous math.
As illustrated in Figure~\ref{fig:intro_idea}, 
the channel with a low class-discriminative score has overlapped 2D projection and shows little class discrepancy visually.
Such redundant channel is pruned away
while the one with rich classification information is kept.

\vspace{-8pt}

However, the effectiveness of the class-discriminative metrics in the prior works is not well studied.
While~\cite{zhuang2018discrimination} inserts a cross-entropy type loss and~\cite{kung2019methodical} uses a closed-form function,
the metrics are proposed and evaluated singly without comparing to other existing well-defined discriminant functions, e.g., Maximum Mean Discrepancy (MMD)~\cite{gretton2012kernel} 
on the channel pruning task.
Moreover, well known single-variate binary-class statistics, like Student's T-Test~\cite{lehmann2006testing}, 
are also mathematically defined to test the significance of two class's difference, 
whilst no attempt has been made to generalize them for high-dimensional multi-label channel scoring.
Moreover,
current practices require iterative optimization steps 
or heavy matrix operation like matrix inversion to obtain the discrimination score,
which can hinder the methods to be applied to larger networks and datasets.

We here have the first attempt to study the effectiveness over a variety of class-discriminative functions on channel pruning.
The group of the interested functions includes not only high-dimensional metrics like the one proposed in~\cite{kung2019methodical} and MMD,
but also several well-known single-variate binary-class statistics like Student's T-Test~\cite{lehmann2006testing}, 
for which we initiate an intuitive and lightweight approach to generalize them for the high-dimensional multi-class channel scoring. 
Surprisingly, we find that one of our generalized single-variate metric, generalized Symmetric Divergence (\metric), outperforms other functions notably,
while having a relatively low computation complexity.
Qualitative and quantitative analyses show 
that \metric\ largely improves the selected channels' quality over prior methods, including non-discriminative type metrics.
In addition,
we propose a \analysis\ scheme to evaluate the layer's pruning sensitivity under the same FLOPs reduction.
This scheme allows us to fairly compare layers' redundancy across the model 
and automatically determine the pruning layers and number of pruning channels,
which further enhances the pruning performance.
We integrate \metric\ and the \analysis\ into a unified pruning algorithm, 
via which we achieve state-of-the-art channel pruning results.

Our contributions are summarized as follows: 
(1) We conduct the first study on the effectiveness of a group of discriminant functions for channel pruning.  
The effectiveness of the winning metric, \metric, is well explained by our qualitative and quantitative channel selection analysis,
where it outperforms state-of-the-art methods with a clear margin. 
(2) We propose a \analysis\ for cross-layer pruning decision.
We develop a pruning algorithm with the scheme and \metric, which allows us to remove redundant channels without performance loss. 
(3) We demonstrate the advantages of our approach on three benchmark datasets (CIFAR-10, CIFAR-100, and ILSVRC-2012) with two widely used networks (VGGNet and ResNet). 
On ILSVRC-2012, our pruned ResNet-50 achieves 75.85\% top-1 accuracy with 44.3\% FLOPs reduction, surpassing the state of the art.

    
    
\section{Related Work}

{\bf Non-pruning Approaches.}
To reduce the computation complexity of CNNs,  early works propose to factorize convolution kernels~\cite{jaderberg2014speeding,lebedev2014speeding, zhang2015efficient, tai2015convolutional}.     
Jaderberg et al.~\cite{jaderberg2014speeding} explore a low-rank basis of filters to represent the full convolutional kernels,
while Lebedev et al.~\cite{lebedev2014speeding} decompose a wide convolutional layer into a sequence of thinner layers. 
In addition, weight quantization methods are studied for CNN reduction~\cite{chen2015compressing, courbariaux2016binarized, han2015deep, rastegari2016xnor, zhou2016dorefa}.
Chen et al.~\cite{chen2015compressing} hash weights to different groups for storage saving. 
Courbariaux et al.~\cite{courbariaux2016binarized} and Rastegari et al.~\cite{rastegari2016xnor} binarize the network's weights for inference speedup. 
Our methods can be applied directly upon these approaches to further accelerate the CNNs without specialized hardware/software.

{\bf Weight-level Pruning.}
Weight-level pruning aims to remove redundant weights, 
resulting in sparsely compressed networks~\cite{hassibi1993second, han2015learning, guo2016dynamic, tung2018clip, carreira2018learning, zhang2018systematic}. 
Han et al.~\cite{han2015learning} trim weights whose magnitudes are lower than a predefined threshold.
Zhang et al.~\cite{zhang2018systematic} adopt the alternating direction method of multipliers~\cite{boyd2011distributed} for weight pruning.
However, the produced sparse networks require specialized software libraries and hardware architectures to achieve the nominal acceleration. 

{\bf Channel-level Pruning.}
In contrast, channel-level pruning can reduce network size and accelerate inference speed without specialized tools~\cite{li2016pruning, liu2017learning, louizos2017learning, he2017channel, luo2017thinet, yu2018nisp, huang2018data, zhuang2018discrimination, he2019filter, kung2019methodical, ye2018rethinking}.
The core idea of channel pruning is to score the importance of channels and then drop the worthless ones. 
The channels' importance can be assessed via their associated filters~\cite{li2016pruning,liu2017learning,louizos2017learning, he2019filter, he2018soft}. 
Li et al.~\cite{li2016pruning} adopt $\ell1$-norm of the filters as the pruning criterion.
He et al.~\cite{he2019filter} calculate the filters' geometric median and discard filters closest to it. 
Scoring channels' saliency based on their activations is another popular direction~\cite{he2017channel, luo2017thinet, huang2018data, yu2018nisp, zhuang2018discrimination, kung2019methodical, lin2019towards}.
He et al.~\cite{he2017channel} and Luo et al.~\cite{luo2017thinet} remove channels that have the least impact on the reconstructed mean square error. 
Yu et al.~\cite{yu2018nisp} propose neural importance score propagation (NISP) to score the channel importance in a more general manner. 

In line with our work, 
several discrimination-based channel scoring methods have improved the efficacy of pruning~\cite{zhuang2018discrimination, kung2019methodical}.
Zhuang et al.~\cite{zhuang2018discrimination} insert discriminant losses to intermediate layers and prune channels that are less correlated to the losses after iterative steps. 
Kung et al.~\cite{kung2019methodical} achieve state-of-the-art pruning results by using a closed-form class-discriminative function, Discriminant Information (DI), for channel selection.
While these studies solely focus on a single discriminant loss/metric,
we provide more analysis on the effectiveness across a wide range of metrics, which is absent in the prior study. 
Moreover, we also pioneer to generalize conventional univariate binary-class statistics to serve for our channel importance evaluation.
The winning metric in our study has much less time complexity
comparing to the time-consuming loss optimization and DI's heavy matrix inversion,
while achieves a noticeable improvement upon them.

\vspace{-0.1cm}
\section{Methodology}
 
\subsection{Discriminant Function for Channel Pruning}\label{sec:metrics}

{\bf Function Effectiveness Study.}
Although discriminant functions are all theoretically convincing, their empirical effectiveness on channel pruning remains understudied. 
To obtain more insights into their practical usefulness, we conduct a study over a group of discriminant functions.
Our study includes high dimensional discrepancy metrics:
Discriminant Information (DI)~\cite{kung2019methodical} and
Maximum Mean Discrepancy (MMD)~\cite{gretton2012kernel}.
We also observe a group of conventional single-variate binary-class metrics which are well-defined for two-class discrimination measurement: 
Student's T-Test (Ttest)~\cite{lehmann2006testing}, 
Absolute SNR (AbsSNR)~\cite{golub1999molecular},
Symmetric Divergence (SD)~\cite{mak2006solution}, 
and Fisher Discriminant Ratio (FDR)~\cite{pavlidis2001gene}.
We provide an intuitive approach to generalize them to high-dimensional multi-class channel scoring
and name the generalized ones with a prefix 'G-' (G-Ttest, \metric, .etc). 
Our implementation of these functions can be found in the Supplementary Material.

\begin{figure*}[t!]
    \centering
    \begin{subfigure}[t]{0.5\textwidth}
        \centering
        \includegraphics[width=\textwidth]{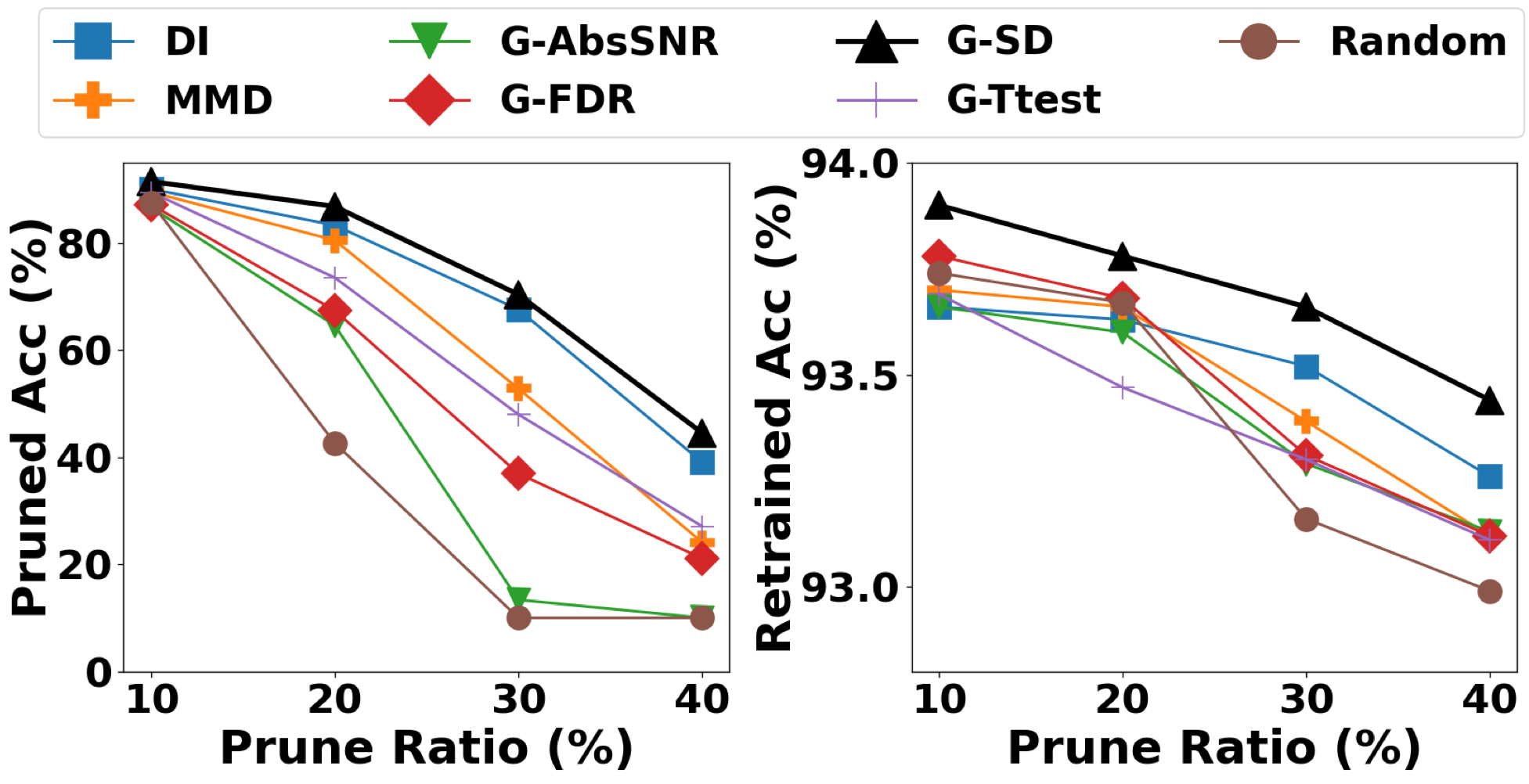}
        \caption{VGG-16 on CIFAR-10}
    \end{subfigure}%
    ~ 
    \begin{subfigure}[t]{0.5\textwidth}
        \centering
        \includegraphics[width=\textwidth]{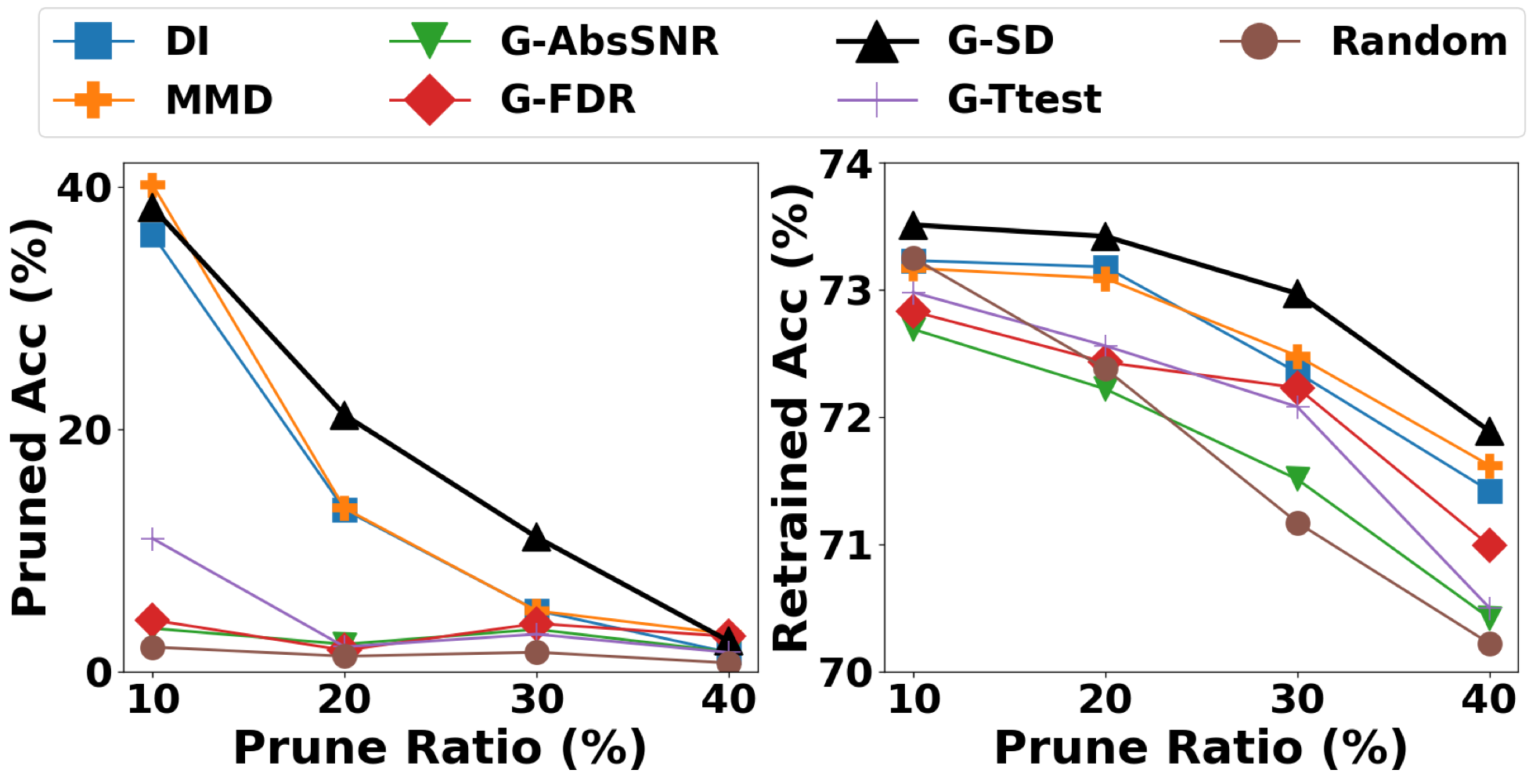}
        \caption{ResNet-38 on CIFAR-100}
    \end{subfigure}
    \caption{Function effectiveness study. 
    The discriminant functions are applied to prune the networks with different layer pruning ratios. 
    The pruned accuracies (without retraining) and retrained accuracies (with retraining) are plotted with respect to layer pruning ratios.}
    \label{fig:function effectiveness study}
\end{figure*}

We study the empirical effectiveness of these functions by using them to conduct one-shot pruning for VGG-16 on CIFAR-10 and ResNet-38 on CIFAR-100,
whose results are shown in Fig.~\ref{fig:function effectiveness study}.
For each metric, we uniformly prune 10\%, 20\%, 30\%, and 40\% of channels with the lowest scores for all layers in both networks. 
These pruned models are then fine-tuned for 150 epochs by the SGD optimizer with Nesterov Momentum. 
All training hyper-parameters are the same for all metrics to allow a fair comparison. 
We evaluate the accuracies of the pruned models before and after the fine-tuning process, 
and include a random scoring method as the baseline. 
We find that the group of discriminant functions outperform the random baseline in nearly all pruning ratios, with and without fine-tuning.
Moreover, we find a subset of leading metrics: \metric, DI, and MMD, where \metric\ has a clear improvement over the others. 
Without retraining, \metric\ outperforms DI by 5.5\% accuracy on CIFAR-10 with 40\% channels removed
and has an 8\% accuracy gain on CIFAR-100 with 30\% channels removed over MMD.
With retraining, \metric\ also performs the best among all other metrics on both tasks.
For instance, it gains around 0.5\% accuracy over MMD on CIFAR-100 when 30\% channels removed.
Based on such consistent winning results, we adopt \metric\ as our channel selection criterion to conduct more class-discriminative pruning experiments.

{\bf Winning Metric.}
Let us denote an $n$-sample $2$-class single-variate dataset as $\mathcal{D} = \{ (x_i,y_i) \}_{i=1}^n$, 
where $y_i \in \{0,1\}$ is the class label associated with each sample $x_i \in \mathbb{R}$. 
Let $\mathcal{P} = \{ x_i~|~(x_i,y_i)\in \mathcal{D},~y_i=1 \}$ and 
$\mathcal{Q} = \{ x_i~|~(x_i,y_i)\in \mathcal{D},~y_i=0 \}$ denote two partitions of $\mathcal{D}$ based on the labels, 
we can then formally define Symmetric Divergence (SD)~\cite{kung2014kernel} of $\mathcal{D}$ as:
\begin{equation}\label{eqn:SD}
   \mathrm{SD}(\mathcal{P}, \mathcal{Q}) = \frac{1}{2} \left( 
\frac{\sigma_P^2}{\sigma_Q^2} + \frac{\sigma_Q^2}{\sigma_P^2}
\right) 
+ 
\frac{1}{2}
\left( 
\frac{ (\mu_P - \mu_Q)^2 }
{\sigma_P^2 + \sigma_Q^2}
 \right) - 1 
\end{equation}
where $\mu_P$ and $\sigma_P^2$ are the sample mean and sample variance of 
$\mathcal{P}$, and $\mu_Q$ and $\sigma_Q^2$ are the corresponding statistics of $\mathcal{Q}$.

SD is initially proposed to select discriminative individual features in bioinformatics feature vectors for dimension reduction and efficient classification~\cite{mak2006solution}. 
The discriminant power of a single scalar feature is quantified by the first two terms of SD.
The first term measures the {\itshape divergence} of two classes {\itshape symmetrically}. 
It achieves a high score for the feature whose $\sigma_P^2$ or $\sigma_Q^2$ is much smaller than the other,
which indicates that one class of the feature has a more concentrated distribution, and thus the feature is more class distinguishable. 
The second term\footnote{The expressions of SD's second term vary a bit in~\cite{kung2014kernel} and~\cite{mak2006solution}.
In this work, we adopt the one in~\cite{kung2014kernel}.
} 
awards features that have large ``Signal-to-Noise Ratio (SNR)'' \cite{jaynes1957information} 
where the centroids distance $(\mu_P - \mu_Q)^2$ can be viewed as ``class discrimination signal" and the class variance sum $\sigma_P^2 + \sigma_Q^2$ is regarded as ``class variational noise''.
Therefore, SD's ability to find discriminant features is mathematically supported. 

To generalize such function for the channel's importance evaluation, 
the key is to find the corresponding statistics, $\mu_P$, $\sigma_P^2$, $\mu_Q$, and $\sigma_Q^2$ for the setting of channel scoring. 
To achieve this, we partition the feature maps of a channel in a one-versus-rest manner.
We then apply a similar statistics operation on the partitioned high-dimensional tensor to obtain their corresponding $\mu_P$, $\sigma_P^2$, $\mu_Q$, and $\sigma_Q^2$.
These statistics are then plugged back to Eqn.~\ref{eqn:SD} to obtain the SD score for such partition.
We finally aggregate the SD score for each partition to obtain a single scalar, \metric\ score, which summarizes the channel's importance. 
The details of such generalization can be found in Supplementary Material.
The effectiveness of \metric\ over state-of-the-art methods are further evidenced in our channel selection analyses 
shown in Fig.~\ref{fig:Qualitative Channel Selection} and Fig.~\ref{fig:Quantitative Channel Selection}.
\metric\ incurs no expensive operations (e.g., matrix inversion, SVD),
which makes it scalable to large networks and datasets.

\subsection{FLOP-Normalized Sensitivity Analysis}

{\bf Pruning Algorithm.}
For deep CNNs, 
layers are constructed heterogeneously as the number of FLOPs,
the number of parameters, and the spatial size 
all vary across layers.
Due to this, selecting pruning layers and determining number of pruning channels are crucial in the pruning procedure.
A method to address the issue is to conduct the layer sensitivity analysis~\cite{li2016pruning}, 
where a two-step approach is proposed to quantitate a layer's sensitivity: 
(1) remove a certain ratio of redundant channels from the layer;
(2) evaluate the validation accuracy of the single-layer-pruned model without retraining. 
Such evaluation is repeated with different pruning ratios, and a pruning ratio vs. accuracy curve can be drawn for each layer to visualize the layer's sensitivity.
This inevitably makes the method labor-intensive
as the pruning layers and the number of pruned channels are decided via human analysis on each layer's curve.
Moreover, in step (1), removing the same ratio of channels in different layers could result in drastically different FLOPs reduction, 
which could make the analysis unfair.

To remedy these issues,
we propose \analysis\ in Algorithm~\ref{al:FLOPs-based Sensitivity Analysis}
to automate the structural pruning process with attention to the normalization of FLOP reduction.
The effectiveness of our approach is shown in Fig.~\ref{fig:FLOPs-based_vs_Normal}.
Let us denote an $L$-layer model by $\mathrm{\bf M}$, we aim to automatically determine the structure of the pruned model $\mathrm{\bf M'}$.
We denote a single-layer-pruned model by $\mathrm{\bf M_\ell^{{\shortminus}n}}$, which indicates that $n$ channels are removed from layer $l$ in $\mathrm{\bf M}$.
For each layer, we remove one single channel and calculate the resultant FLOPs reduction of the network by Eqn.~\ref{eqn:FLOSS}. 
Such single-channel FLOPs loss at the $l$-th layer is denoted as $\mathrm{FLOSS_\ell}$. 
\vspace{-0.05cm}
\begin{equation}
\label{eqn:FLOSS}
\mathrm{FLOSS_\ell = FLOP(\mathrm{\bf M}) - FLOP(\mathrm{\bf M_\ell^{{\shortminus}1}}}) 
\end{equation}
The number of pruning channels from layer $l$, $n_l$, can thus be determined in Eqn.~\ref{eqn:num channel removal},
where $\mathrm{FLOSS_{max}} = \max_{\ell \in [1:L]} \mathrm{FLOSS_\ell}$ and $\alpha$ is a scaling factor to control the pruning rate. 
Via the normalization of $\mathrm{FLOSS_{max}} / \mathrm{FLOSS_\ell}$, removing $n_l$ channels from layer $l$ would have the same overall network's FLOPs reduction. The derived $n_l$ will be further rounded to integer.
\vspace{-0.05cm}
\begin{equation}
\label{eqn:num channel removal}
n_\ell = \mathrm{round}(\alpha \times \mathrm{FLOSS_{max}} / \mathrm{FLOSS_\ell})
\end{equation}
We then follow the two-step anlaysis scheme: 
(1) individually prune $n_l$ channels with the least \metric\ from layer $l$ to obtain $\mathrm{\bf M_\ell^{{\shortminus}n_\ell}}$;
(2) test the validation accuracy of $\mathrm{\bf M_\ell^{{\shortminus}n_\ell}}$, $\mathrm{Acc}_l$. 
$\mathrm{Acc}_l$ effectively quantifies layer $l$'s sensitivity at  the same FLOP's reduction level.
Finally, the pruned network $\mathrm{\bf M'}$ is obtained by choosing $k$ layers with the highest $\mathrm{Acc}_l$ and pruned associated $n_l$ channels from these layers.
After structural pruning, $\mathrm{\bf M'}$ will be fine-tuned to recover the accuracy loss. 
By iterating this pruning-retraining process, we can obtain a compact model with high performance.

\begin{minipage}[t]{0.63\textwidth}
      \begin{algorithm}[H]
        {\footnotesize
    \SetKwInOut{Input}{Input}
    \SetKwInOut{Output}{Output}
    \Input{Model $\mathrm{\bf M}$, Number of Pruning Layers $k$}
    \Output{Pruned Model $\mathrm{\bf M'}$}
	
    Calculate $\mathrm{FLOSS_\ell}$ by Eqn.~\ref{eqn:FLOSS}
	
    Calculate number of pruning channels $n_l$ by Eqn.~\ref{eqn:num channel removal}

{\bf Evaluate Layer's Sensitivity:} \For{ $l \in  [1:L]$ }{
Obtain $\mathrm{\bf M_\ell^{{\shortminus}n_\ell}}$ by temporarily removing $n_l$ channels in layer $l$

Test validation accuracy of $\mathrm{\bf M_\ell^{{\shortminus}n_\ell}}$, $\mathrm{Acc}_l$
}

Select layers with top-k $\mathrm{Acc}$, $\{l_1, \cdots,l_k\}$, and their number of pruning channels$\{n_1, \cdots, n_k\}$

{\bf Obtain Pruned Structure:} \For{ $i \in  [1:k]$ }{
Permanently prune $n_i$ channels in layer $l_i$ from $\mathrm{\bf M}$
}
    return $\mathrm{\bf M'}$
}
    \caption{\label{al:FLOPs-based Sensitivity Analysis} FLOP-Normalized Sensitivity Analysis}
\end{algorithm}
\end{minipage}
\hfill
\begin{minipage}[t]{0.33\textwidth}
{\bf Handling Skip Connection.}
For CNNs like ResNet, the input of a residual block is added to its output,
which requires the number of channels to be the same for the block's input and output.
We adopt the idea of using a channel selection layer at the beginning of a residual block~\cite{he2017channel,liu2017learning}. 
Such layer samples a subset of input channels for the first convolution layer of the block based on their discriminative scores,
which further reduces the block's FLOPs and parameters without overheads.
\end{minipage}

       







  


  



\section{Experimental Results}\label{experiment}

{\bf Benchmarks.} We empirically evaluate our \metric\ pruning method with VGGNet~\cite{simonyan2014very} and ResNet~\cite{he2016deep}
on three image classification datasets,
CIFAR-10~\cite{krizhevsky2009learning}, CIFAR-100~\cite{krizhevsky2009learning}, and ILSVRC-2012~\cite{deng2009imagenet}. 
CIFAR-10/100 contains 50K training samples and 10K test samples from 10/100 classes. 
ILSVRC-2012 contains 1.28M training images and 50K validation images in 1000 classes. 
We report pruned models at different FLOPs levels by adding a letter suffix (e.g., \metric-A and \metric-B).
We compare \metric\ pruning with previous approaches, e.g.,
ThiNet~\cite{luo2017thinet}, 
DCP~\cite{zhuang2018discrimination},
GAL~\cite{lin2019towards},
FPGM~\cite{he2019filter}, 
KSE~\cite{li2019exploiting},
TAS~\cite{dong2019network},
and DI~\cite{kung2019methodical}.
Our method achieves state-of-the-art results on all benchmarks.

{\bf Training.}
Experiments are implemented in TensorFlow~\cite{abadi2016tensorflow} 
and are carried out with NVIDIA Tesla P100 GPUs. 
We use the SGD optimizer with Nesterov Momentum~\cite{nesterov1983method} with a momentum of 0.9.
The weight decay factor is set to 0.0001.
We use the standard data augmentation scheme proposed in~\cite{he2016deep} for all datasets. 
On CIFAR, we fine-tune 200 epochs with a minibatch size of 128. 
The learning rate is initialized at 0.01 and divided by 5 at epoch 80 and 160. 
On ILSVRC-2012, we use a batch size of 256 to fine-tune VGG-16/ResNet-50 with 30/120 epochs.
The learning rate is started at 0.0003 and multiplied by 0.4 at 40\% and 80\% of the total number of epochs.
During the evaluation, the images are resized and center-cropped to the size of 224$\times$224.

\begin{table}[H]
\begin{center}
\fontsize{8.5}{10}\selectfont

(a) CIFAR-10 Dataset

\begin{tabular}{|c|c|c|c|c|c|c|c|c|}
    \hline
    Network & Method & Test Acc. (\%) & Acc. $\downarrow$ (\%) & FLOPs & Pruned (\%) & Parameters & Pruned (\%) \\
    \hline\hline
    

    \multirow{4}{3em}{VGG16} & L1 \cite{li2016pruning} & 93.25 $\rightarrow$ 93.40 & -0.15 & 211M & 34.2 & 5.40M & 64.0 \\
    
    & GAL \cite{lin2019towards} & 93.96 $\rightarrow$ 93.42 & 0.54 & 172M & 45.2 & 2.67M & 82.2 \\
    
    & SSS \cite{huang2018data} & 93.96 $\rightarrow$ 93.02 & 0.94 & 183M & 39.6 & 3.93M & 73.8 \\

    & {\bf \metric} & {\bf 93.45 $\rightarrow$ 93.68} & {\bf -0.23} & {\bf 62M} & {\bf 80.1} & {\bf 0.57M} & {\bf 96.2} \\ 

    \hline\hline
    
    
    \multirow{9}{3em}{\thead{ResNet \\ 56}}
    & L1 \cite{li2016pruning} & 93.04 $\rightarrow$ 93.06 & -0.02 & 91M & 27.6 & 730K & 13.7 \\

    & GAL \cite{lin2019towards} & 93.26 $\rightarrow$ 93.38 & -0.12 & 78M & 37.6 & 750K & 11.8 \\
    
    & NISP \cite{yu2018nisp} & 93.04 $\rightarrow$ 93.01 & 0.03 & 71M & 43.6 & 500K & 42.6 \\
    
    & DCP \cite{zhuang2018discrimination} & 93.80 $\rightarrow$ 93.49 & 0.31 &  63M & 49.8 & 435K & 49.2 \\
    
    & {\bf \metric-A} & {\bf 93.40 $\rightarrow$ 93.95} & {\bf -0.55} & {\bf 63M} & {\bf 49.6} & {\bf 510K} & {\bf 40.4}
    \\ \cdashline{2-8}

    & TAS \cite{dong2019network} & 94.46 $\rightarrow$ 93.69 & 0.77 & 60M & 52.7 & - & - \\
    
    & FPGM \cite{he2019filter} & 93.59 $\rightarrow$ 93.49 & 0.10 & 59M & 52.6 & - & -\\
    
    & KSE \cite{li2019exploiting} & 93.03 $\rightarrow$ 92.88 & 0.15 & 50M & 60.0 & 306K & 58.3 \\    
    
    & {\bf \metric-B} & {\bf 93.40 $\rightarrow$ 93.84} & {\bf -0.44} & {\bf 46M} & {\bf 63.0} & {\bf 407K} & {\bf 52.4} \\ 
    \hline\hline
    

    \multirow{6}{3em}{\thead{ResNet \\ 110}} & L1 \cite{li2016pruning} & 93.53 $\rightarrow$ 93.30 & 0.23 & 155M & 38.6 & 1.18M & 32.2  \\
    
    & NISP \cite{yu2018nisp} & 93.53 $\rightarrow$ 93.38 & 0.15 & 142M & 43.8 & 0.99M & 43.3 \\ 
    
    & {\bf \metric-A} & {\bf 93.70 $\rightarrow$ 94.45} & {\bf -0.75} & {\bf 136M} & {\bf 46.1} & {\bf 1.01M} & {\bf 41.6} \\ 
    \cdashline{2-8}

    & GAL \cite{lin2019towards} & 93.50 $\rightarrow$ 92.74 & 0.76 & 130M & 48.5 & 0.95M & 44.8 \\
    
    & FPGM \cite{he2019filter} & 93.68 $\rightarrow$ 93.74 & -0.06 & 121M & 52.3 & - & -\\
    
    & {\bf \metric-B} & {\bf 93.70 $\rightarrow$ 94.04} & {\bf -0.34} & {\bf 101M} & {\bf 60.0} & {\bf 0.70M} & {\bf 59.4} \\

    \hline
    \end{tabular}

\medskip

(b) CIFAR-100 Dataset

 \begin{tabular}{|c|c|c|c|c|c|c|c|c|}
    \hline
    Network & Method & Test Acc. (\%) & Acc. $\downarrow$ (\%) & FLOPs & Pruned (\%) & Parameters & Pruned (\%) \\
    \hline\hline
    
    
    \multirow{2}{3em}{VGG19} & SLIM~\cite{liu2017learning} & 73.26 $\rightarrow$ 73.48 & -0.22 & 256M & 37.1 & 5.0M & 75.1 \\
    
    & {\bf \metric} & {\bf 73.40 $\rightarrow$ 73.67} & {\bf -0.27} & {\bf 161M} & {\bf 59.5} & {\bf 3.2M} & {\bf 84.0} \\ 

    \hline\hline
    
    
    \multirow{4}{3em}{\thead{ResNet \\ 164}} & LCCN \cite{dong2017more} & 75.67 $\rightarrow$ 75.26 & 0.41 & 197M & 21.3 & - & - \\
    
    & SLIM~\cite{liu2017learning} & 76.63 $\rightarrow$ 76.09 & 0.54 & 124M & 50.6 & 1.21M & 29.7\\
    
    & DI~\cite{kung2019methodical} & 76.63 $\rightarrow$ 76.11 & 0.52 & 105M & 58.0 & 0.95M & 45.1\\

    & {\bf \metric} & {\bf 76.95 $\rightarrow$ 77.40} & {\bf -0.45} & {\bf 92M} & {\bf 63.2} & {\bf 0.66M} & {\bf 61.8} \\
    
    \hline
    \end{tabular}

\end{center}
\caption{Experiments on CIFAR dataset. `-': Results not reported in original papers.}\label{tab:CIFAR_Experiments}
\end{table}
\vspace{-0.55cm}
{\bf Pruning.} On CIFAR, all training data are feed-forwarded for \metric\ channel scoring. 
On ILSVRC-2012, we randomly sample 10,000 training images for \metric\ calculation in each pruning iteration.
We choose $k$ to be around one-third of the total number of layers and $\alpha \in [2,4]$ for Algorithm~\ref{al:FLOPs-based Sensitivity Analysis}.

\subsection{CIFAR Experiments}
We examine the performance of our scheme on CIFAR 
with five popular networks, 
VGG-16,19 and ResNet-56,110,164. 
We use the same VGGNet structures as~\cite{li2016pruning,liu2017learning} 
with slim fully connected layers.
As shown in Table~\ref{tab:CIFAR_Experiments}, we achieve state-of-the-art results on both datasets for all networks.

On CIFAR-10, 
our VGG-16 has 5$\times$ storage saving and 3$\times$ inference speedup,  
while having 0.26\% higher accuracy compared to GAL~\cite{lin2019towards}. 
Our \metric-A of ResNet-56 outperforms DCP~\cite{zhuang2018discrimination}
with 0.46\% accuracy gain at the same FLOPs reduction ratio.
Compared to FPGM~\cite{he2019filter} on ResNet-110, 
\metric-B achieves 94.04\% accuracy, which is 0.3\% higher while having 8\% less FLOPs. 
On CIFAR-100,
our pruned VGG-19 has an accuracy gain of around 0.2\% comparing to SLIM~\cite{liu2017learning} 
with over 20\% less FLOPs and around 10\% fewer parameters. 
On ResNet-164, 
our pruned model outperforms DI~\cite{kung2019methodical} by 1.29\% in accuracy 
with 5\% less FLOPs and 16\% fewer parameters.
These results demonstrate that \metric\ can produce more compact models with better performance 
compared to prior methods.

\subsection{ILSVRC-2012 Experiments}

We further conduct experiments on VGG-16 and ResNet-50 for large-scale ImageNet classification tasks.
The results are summarized in Table~\ref{tab:ImageNet_Experiments} and we achieve state-of-the-art performance, again.

On VGG-16, 
our \metric-A increases the performance in both top-1 and top-5 accuracy by 0.58\% and 0.56\% 
with 2.3$\times$ acceleration from baseline.
This result shows that the proposed metric has a strong regularization effect on deep CNNs,
which helps remove deleterious nodes and improves 
generalization performance.
Compared to FBS~\cite{gao2018dynamic},  our 3.3$\times$ accelerated \metric-B 
owns a higher speedup 
and compression ratio, 
while exceeds by 0.50\% in top-5 accuracy.
On ResNet-50, \metric-A 
\begin{table}[H]
\begin{center}

\fontsize{8.5}{10}\selectfont

\begin{tabular}{|c|c|c|c|c|c|c|c|}
    \hline
    Network & Method & \thead{ Top-1 \\ Acc. (\%)}  & \thead{ Top-1 \\ $\downarrow$ (\%)} & \thead{Top-5 \\ Acc. (\%)} & \thead{Top-5 \\ $\downarrow$ (\%)} & \thead{FLOPs (B) \\ Pruned (\%)}  & \thead{Params (M) \\ Pruned (\%)} \\
    \hline\hline
    
    
    
    \multirow{7}{3.5em}{VGG16} & L1 \cite{li2016pruning} & - & - & 89.90 $\rightarrow$ 89.10 & 0.80 & 7.74 (50.0) & -  
    \\
    
    & CP \cite{he2017channel} & - & - &
    89.90 $\rightarrow$ 89.90 & 0.00 & 7.74 (50.0) & -  
    \\
    
    & {\bf \metric-A} & {\bf 71.30 $\rightarrow$ 71.88} & {\bf -0.58} & {\bf 90.10 $\rightarrow$ 90.66} & {\bf -0.56} & {\bf 6.62 (57.2)} & {\bf 133.61 (3.4)} \\ \cdashline{2-8}

    & RNP \cite{lin2017runtime} & - & - & 89.90 $\rightarrow$ 86.67 & 3.23 & 5.16 (66.7) & 138.34 (0.0) \\
    
    & SLIM \cite{liu2017learning} & - & - & 89.90 $\rightarrow$ 88.53 & 1.37 & 5.16 (66.7) & - 
    \\
    
    & FBS \cite{gao2018dynamic} & - & - &
    89.90 $\rightarrow$ 89.86 & 0.04 & 5.16 (66.7) & 138.34 (0.0) 
    \\
    
    & {\bf \metric-B} & {\bf 71.30 $\rightarrow$ 71.26} & {\bf 0.04} & {\bf 90.10 $\rightarrow$ 90.36} & {\bf -0.26} & {\bf 4.68 (69.7)} & {\bf 131.19 (5.2)} \\ 
    \hline\hline


    \multirow{9}{3.5em}{\thead{ResNet \\ 50}}& SSS \cite{huang2018data} & 76.12 $\rightarrow$ 74.18 & 1.94 & 92.86 $\rightarrow$ 91.91 & 0.95 & 2.82 (31.0) & - \\
    
    & L1 \cite{li2016pruning} & 76.12 $\rightarrow$ 72.88 & 3.24 & 92.86 $\rightarrow$ 91.05 & 1.81 & 3.07 (24.9) & - \\

    & {\bf \metric-A} & {\bf 76.15 $\rightarrow$ 76.21} & {\bf -0.06} & {\bf 92.87 $\rightarrow$ 92.92} & {\bf -0.05} & {\bf 2.90 (29.1)} & {\bf 21.91 (14.1)} \\ 
    \cdashline{2-8}
    
    & ThiNet \cite{luo2017thinet} & 72.88 $\rightarrow$ 72.04 & 0.84 & 91.14 $\rightarrow$ 90.67 & 0.47 & 2.44 (40.3) & 16.94 (33.6) \\
    
    & NISP \cite{yu2018nisp} & 72.88 $\rightarrow$ 71.99 & 0.89 & - & - & 2.29 (44.0) & 14.36 (43.7) \\
    
    & GAL \cite{lin2019towards} & 76.15 $\rightarrow$ 71.95 & 4.20 & 92.87 $\rightarrow$ 90.94 & 1.97 & 2.33 (43.0) & 21.20 (16.9) \\
    
    & SFP \cite{he2018soft} & 76.15 $\rightarrow$ 74.61 & 1.54 & 92.87 $\rightarrow$ 92.06 & 0.81 & 2.38 (41.8) & - \\
    
    & {\bf \metric-B} & {\bf 76.15 $\rightarrow$ 75.85} & {\bf 0.30} & {\bf 92.87 $\rightarrow$ 92.66} & {\bf 0.21} & {\bf 2.28 (44.3)} & {\bf 19.59 (23.2)} \\ 
    


    \hline

\end{tabular}

\end{center}
\caption{
Experiments on ILSVRC-2012 dataset. `-': Results not reported in original papers.}
\label{tab:ImageNet_Experiments}
\end{table}












\vspace{-0.6cm}
increases top-1 and top-5 accuracy both by around 0.05\% at 30\% of FLOPs reduction,
which is rarely reported
on ILSVRC-2012.
As ResNet-50 has much less redundancy than VGG-16, such accuracy gain is even more promising.
Moreover, \metric-B achieves a 44.3\% of FLOPs reduction with just 0.30\% top-1 accuracy loss, which surpasses all previous methods.
Compared to GAL~\cite{lin2019towards}, 
\metric-B has a higher speedup ratio while achieves a nearly 4\% top-1 accuracy gain from the same baseline.

\section{Ablation Study} 

Our noticeable pruning performance is directly related to the use of \metric\ and \analysis. 
We conduct the following ablation study to further illustrate their effectiveness in comparison with state-of-the-art methods.  

\subsection{Qualitative Channel Selection}\label{sec:qualitative_channel_sel}

\begin{figure}[t!]
	\centering
    \includegraphics[width = 0.9\textwidth]{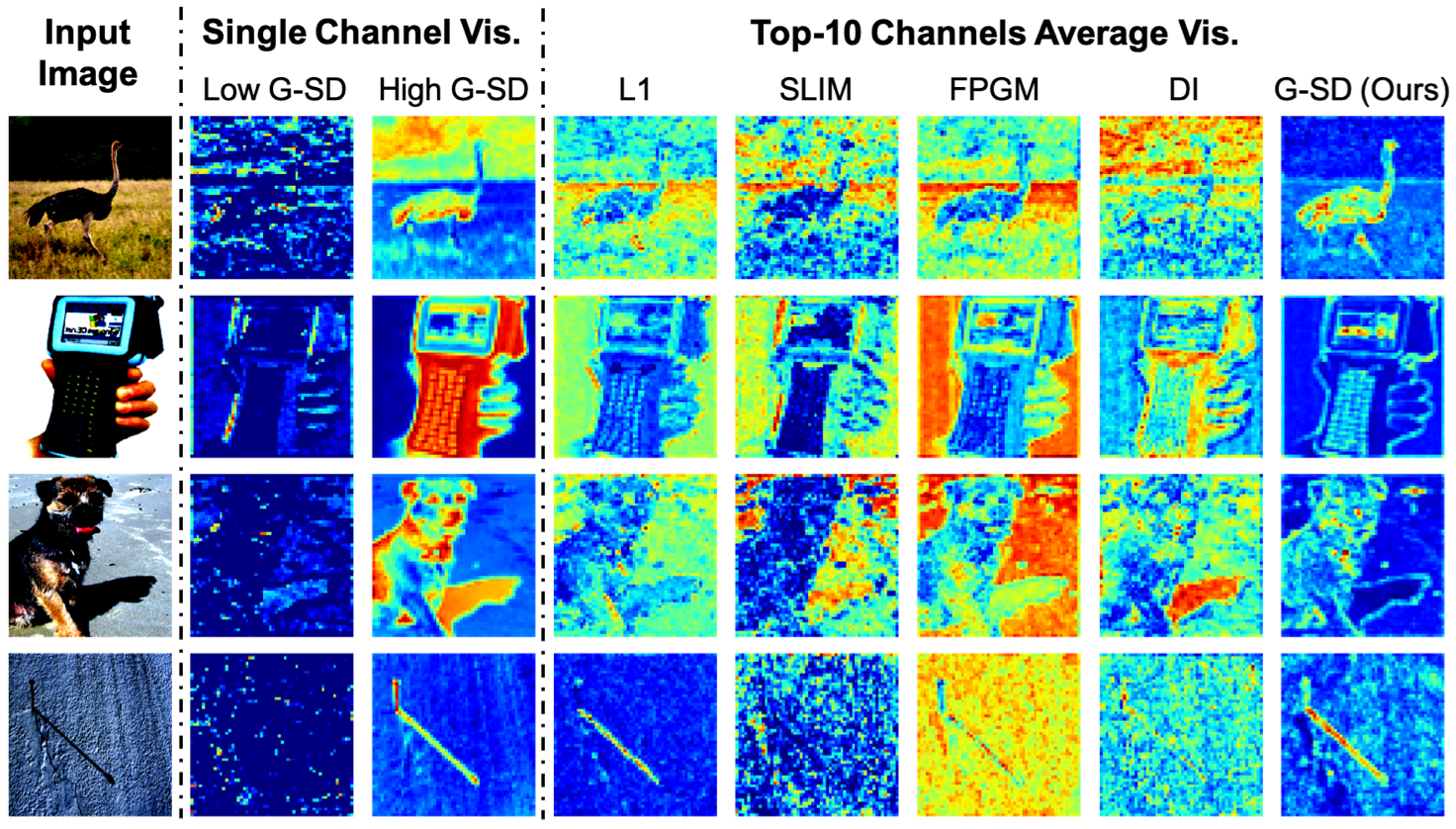}
    \caption{Qualitative channel selection analysis. 
    {\bf Col. 1:} Input images. 
    {\bf Col. 2-3}: Single channel visualization on channels with low and high \metric\ values. 
    {\bf Col. 4-8}: Average responses of the top-10 channels selected by different metrics. From left to right, the metrics are: $\ell$1-norm~\cite{li2016pruning}, batch-norm scaling factor~\cite{liu2017learning}, filter's geometric median~\cite{he2019filter}, DI~\cite{kung2019methodical}, and \metric.
    }
    \label{fig:Qualitative Channel Selection}
\end{figure}

{\bf Single Channel Visualization.}
We visualize the high \metric\ and the low \metric\ channels at the second layer of the first block in ResNet-50 with ILSVRC-2012 inputs.
As shown in Col. 1-3 of Fig.~\ref{fig:Qualitative Channel Selection}, 
we observe that the channel with low \metric\ value ({\bf Col. 2}) tends to generate indistinguishable responses for different classes,
while the high \metric\ channel ({\bf Col. 3}) well preserves the image patterns, which is more informative for classification. 
These suggest that \metric\ is promising for channel selection.

{\bf Multiple Channels Visualization.}
To further visually compare the effectiveness of our \metric\ with other channel selection criteria~\cite{li2016pruning,liu2017learning,he2019filter,kung2019methodical}, 
we score the channel activations 
and compute an average response over the top-10 channels with highest scores for each metric as shown in Col. 4-8 in Fig.~\ref{fig:Qualitative Channel Selection}. 
We observe that the average responses picked by \metric\ ({\bf Col. 8}) 
tends to display more class information than the ones from the other metrics ({\bf Col. 4-7}).
For example, in the first row, \metric\ clearly separates the ostrich from the background grass, while others tend to generate mixed responses.
Moreover, \metric\ is the only one that preserves both the vertical nail and its long diagonal shadow in the fourth row.
Such visual comparison speaks well on behalf of \metric\ effectiveness.

\subsection{Quantitative Channel Selection}

\begin{figure}[t]
    \centering
    \begin{minipage}{0.5\textwidth}
        \centering
        \includegraphics[width=\textwidth, trim={0.24cm 0 0 0}, clip]{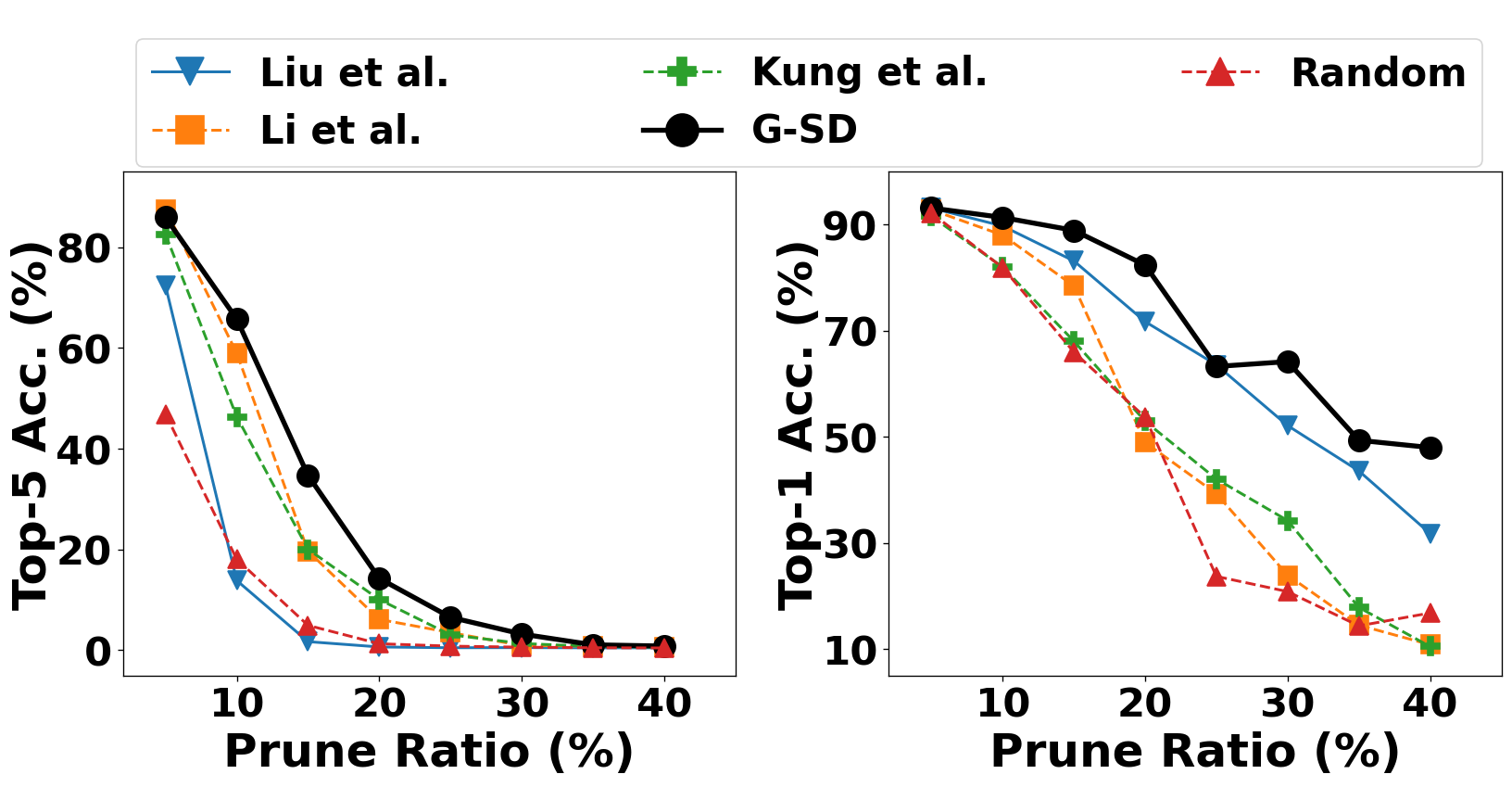}
        \caption{
        The curves of layer pruning ratio vs. accuracy (without retraining) with different channel scoring metrics. 
        {\bf Left:} VGG-16 on ILSVRC-2012.
        {\bf Right:} ResNet-110 on CIFAR-10.
        }
        \label{fig:Quantitative Channel Selection}
    \end{minipage}
    \hfill
    \begin{minipage}{0.45\textwidth}
        \centering
        \includegraphics[width=\textwidth, trim={0.4 0 0 0}, clip]{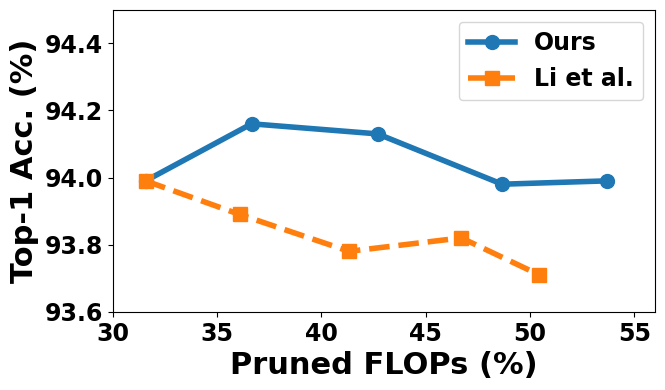}
        \caption{The curves of FLOPs pruning ratio vs. accuracy for two sensitivity analysis methods for ResNet-56 on CIFAR-10.
        }
        \label{fig:FLOPs-based_vs_Normal}
    \end{minipage}
\end{figure}

We demonstrate the advantage of \metric\ over state-of-the-art methods~\cite{li2016pruning,liu2017learning,kung2019methodical} quantitatively in Fig.~\ref{fig:Quantitative Channel Selection},
with VGG-16 on ILSVRC-2012 and ResNet-110 on CIFAR-10.
Similar to the effectiveness study in~\ref{sec:metrics}, 
we use each metric to uniformly remove $r\%$ of the lowest-scored channels in all layers 
and include random scoring as the baseline.  
The accuracy of the pruned networks is evaluated without retraining
to serve as the indicator of channels' quality and the delegate of fine-tuned accuracy~\cite{han2015learning,he2018amc}. 
 $r$ is set to be: $5, 10, 15, 20, 25, 30, 35, 40$.
We observe that \metric\ wins against other methods in nearly all FLOPs reduction ratios on both networks and datasets. 
This analysis well explains \metric's effectiveness by 
comparing the selected channel's quality directly
and eliminating the effect of different retraining settings (optimizer, number of epochs, etc.) across literature.

\subsection{FLOP-Normalized Sensitivity Analysis}

We compare our \analysis\ with the conventional sensitivity analysis proposed in~\cite{li2016pruning} 
with ResNet-56 on CIFAR-10. 
Starting with a 31\% FLOPs-pruned ResNet-56, we conduct five pruning-retraining iterations with \metric\ by two methods. 
As shown in Fig.~\ref{fig:FLOPs-based_vs_Normal}, we find a clear advantage for our methods over Li's in every iteration.
Moreover, the final network obtained by Li's method suffers an accuracy drop of 0.3\% while our model does not have accuracy loss.

    
    
        
    
    
        
    

\vspace{-0.1cm}
\section{Conclusion} 

In this paper, 
we conduct the first study on the empirical effectiveness of a broad range of class-discriminative functions in channel pruning. 
We provide an intuitive generalization to enable single-variate binary-class functions for channel scoring.
We find that the winning metric in our study, generalized Symmetric Divergence (\metric), selects channels with more information over prior arts via qualitative and quantitative analysis.
In addition, we develop a \analysis\ scheme to enhance our pruning performance. 
Experimental results on three benchmark datasets demonstrate the advantage of our mechanism over state-of-the-art methods. 
In the future, we will explore the relationship between the channel's class-discrimination and its redundancy in a more theoretical manner to help design better metrics.

\bibliographystyle{Style_Ref/ieee_fullname}
\bibliography{Style_Ref/refs}

\newpage
\newcommand\feamap{\mathcal{F}}

{\bf \Large Supplementary Materials}

We organize our supplementary material as follows. 
In Section~\ref{sec:DF_study}, we present the definitions of the studied discriminant functions (except SD) 
as well as the details of our generalization for single-variate binary-class functions to high-dimensional multi-class channel scoring.  
In Section~\ref{sec:exp_details}, we discuss the detailed structures of the pruned VGG-16 and ResNet-50 models on ILSVRC-2012 by generalized Symmetric Divergence (\metric). 
We discuss an empirical study on the time complexity of two discriminant functions, 
\metric\ and Discriminant Information (DI)~\cite{kung2019methodical}, for channel scoring in Section~\ref{sec:time_comp}.
In Section~\ref{sec:visual}, we provide more visualizations of 
our qualitative channel selection analysis.
\section{Discriminant Functions}\label{sec:DF_study}

\subsection{Single-Variate Binary-Class Metrics}

We adopt the same notation of the $n$-sample binary-class single-variate dataset as in the main paper.
Our implementation of Absolute SNR (AbsSNR)~\cite{golub1999molecular},
Fisher Discriminant Ratio (FDR)~\cite{pavlidis2001gene} and Student's T-Test (Ttest)~\cite{lehmann2006testing} are:

\begin{equation}
\mathrm{AbsSNR}(\mathcal{P}, \mathcal{Q}) = \frac{|\mu_P - \mu_Q|}{\sigma_P + \sigma_Q}
\end{equation}
\begin{equation}
\mathrm{FDR}(\mathcal{P}, \mathcal{Q})  = \frac{(\mu_P - \mu_Q)^2}{\sigma_P^2 + \sigma_Q^2}
\end{equation}
\begin{equation}\label{eqn:Ttest}
\mathrm{Ttest}(\mathcal{P}, \mathcal{Q})= \frac{|\mu_P - \mu_Q|}{\sqrt{
\frac{\sigma_P^2}{|\mathcal{P}|}
+ \frac{\sigma_Q^2}{|\mathcal{Q}|}
}
}
\end{equation}
where $|\mathcal{P}|$ and $|\mathcal{Q}|$ denote the number of samples in $\mathcal{P}$ and $\mathcal{Q}$.

\subsection{Generalization}\label{sec:generalization}

We observe that the computation of these three metrics along with SD mainly requires four statistics: $\mu_P$, $\sigma_P^2$, $\mu_Q$, and $\sigma_Q^2$.
We here use SD as an example to demonstrate our way to 
find these statistics for CNN's channel scoring, 
while the other three metrics can follow the similar way. 

{\bf Notation.}
For an $N$-sample $C$-class dataset, 
we denote the feature maps of a CNN channel as $\mathcal{F} = \{(f_i, c_i)\}^N_{i = 1}$, 
where $f_{i} \in \mathbb{R}^{W \times H}$ denotes the feature map of the $i$-th input image, 
$c_i \in [1:C]$ is the class label of the $i$-th input image,
and $W/H$ are the spatial sizes of the maps.

{\bf Method.}
We first partition $\feamap$ as $\feamap^c$ and  $\feamap^{-c}$ where 
$\feamap^c = \{f_{i}~|~(f_{i}, c_i) \in \feamap,~c_i = c \}$ and 
$\feamap^{-c} = \{f_{i}~|~(f_{i}, c_i) \in \feamap,~c_i \neq c \}$, $\forall~c \in [1:C] $.  
By this partition, we can find the corresponding statistics of $\mathcal{F} $ in a two-class manner. 
Note each $f_{i}$ in $\feamap^c$ is a 2D feature map with $W \times H$ activations, 
and thus there are $|\feamap^c| \times W \times H$ activations in $\feamap^c$ in total. 
We then define two statistics operators $g_{mean}$ and $g_{var}$ on $\feamap^c$, which return the mean and variance over these $|\feamap^c| \times W \times H$ activations.
We can thus obtain the statistics $\mu_c = g_{mean}(\feamap^c )$ and $\sigma_c^2= g_{var}(\feamap^c )$ for $\feamap^c$.
The counterparts of $\feamap^{-c}$, $\mu_{-c}$ and $\sigma_{-c}^2$, can be computed in the same way. 
The SD score for the two-class partition, ($\feamap^c$ , $\feamap^{-c}$), can thus be derived as:
\begin{equation}\label{eqn:l_class_sd}
   \mathrm{SD}(\feamap^c , \feamap^{-c}) = \frac{1}{2} \left( 
\frac{\sigma_c^2}{\sigma_{-c}^2} + \frac{\sigma_{-c}^2}{\sigma_c^2}
\right) 
+ 
\frac{1}{2}
\left( 
\frac{ (\mu_c - \mu_{-c})^2 }
{\sigma_c^2 + \sigma_{-c}^2}
 \right) - 1 
\end{equation}

$\mathrm{SD}(\feamap^c , \feamap^{-c})$ captures the discriminativenesss of class $c$ relative to the other classes in $\feamap$.
In general, we want to select channels that distinguish all classes well on average.
Thus, under the $C$-class setting, the generalized Symmetric Divergence (\metric) of $\feamap$ is formally defined as:

\begin{equation}\label{eqn:SymDiv}
   \mathrm{G{\shortminus}SD}(\feamap) = \frac{1}{C}\sum_{c = 1}^C {\mathrm{SD}(\feamap^c , \feamap^{-c})}
\end{equation}

\subsection{High-Dimensional Metrics}

We present the implementation of Discriminant Information (DI)~\cite{kung2019methodical} and Maximum Mean Discrepancy (MMD)~\cite{gretton2012kernel} here.
We use the same notation as in Section~\ref{sec:generalization} and regard $f_i$ as a flattened 1D vector.

{\bf DI.} We first define two scatter matrices, $\mathbf{\bar{S}}$ and $\mathbf{S_B}$:
\begin{equation}
\mathbf{\bar{S}} = \sum_{i=1}^N{(f_i - \bar{f})(f_i - \bar{f})^T}
\end{equation}
\begin{equation}
\mathbf{S_B} = \sum_{c=1}^C{|\feamap^c|(\bar{f^c} - \bar{f})(\bar{f^c} - \bar{f})^T}
\end{equation}
where $\bar{f} = \frac{1}{N}\sum_{i=1}^N{f_i}$ and $\bar{f^c} = \frac{1}{N_c}\sum_{i=1}^{N_c}{f_i}$ represents the centroids of the all data and class $c$ ($c \in [1:C]$), respectively. 
DI is then defined as:
\begin{equation}
\mathrm{DI}(\mathcal{F}) = tr([\mathbf{\bar{S}} + \rho\mathbf{I}]^{-1} \mathbf{S_B})
\end{equation}
where $\rho$ is a small ridge factor to ensure invertibility. 
In practice, we set $\rho = 0.0001$.

{\bf MMD.} We define the rbf kernel as:
\begin{equation}
k(x,y) = exp\{-\frac{
||x - y ||^2_2}
{2 \sigma^2}\}
\end{equation}
The two-class MMD on a $(\feamap^c, \feamap^{-c})$ partition can thus be defined as:

\begin{align}
\nonumber\mathrm{MMD}_{two}(\feamap^c, \feamap^{-c}) =~~& \frac{1}{|\feamap^c|^2}\sum_{f_i,f_j\in\feamap^c}{k(f_i, f_j)} 
+\frac{1}{|\feamap^{-c}|^2}\sum_{f_i,f_j\in\feamap^{-c}}{k(f_i, f_j)} \\ &-\frac{2}{|\feamap^{c}|\times|\feamap^{-c}|}\sum_{f_i\in\feamap^c, f_j\in\feamap^{-c}}{k(f_i, f_j)} 
\end{align}

following the similar one-versus rest setting, the MMD score of a channel can be defined as:
\begin{equation}
\mathrm{MMD}(\feamap) = \frac{1}{C}\sum_{c = 1}^C {\mathrm{MMD}_{two}(\feamap^c , \feamap^{-c})}
\end{equation}
In practice, we set $\sigma = 1$ for the rbf kernel.

\section{Pruned Models Details}\label{sec:exp_details}

We show the detailed structures of our pruned VGG-16 and ResNet-50 on ILSVRC-2012 in Table~\ref{tab:VGG-16-Model} and Table~\ref{tab:ResNet-50-Model}. 
Both pruned models are named as \metric-B in the experimental results section.
On VGG-16, we find that our algorithm prunes more channels in shallow layers, 
which indicates the shallow layers are less sensitive to pruning compared to the deep ones under the same overall FLOPs reduction. 
On ResNet-50, our model preserves more channels at the starting blocks of each stage. 
These blocks are the ones where the stride-2 downsamplings happen,
which means that preserving information for the process of resolution reduction would be crucial to maintain the network's performance.

\section{Time Complexity}\label{sec:time_comp}

\begin{figure}[H]
    \centering
    \includegraphics[width=0.5\textwidth, trim={0cm 0 0.1cm 0}, clip]{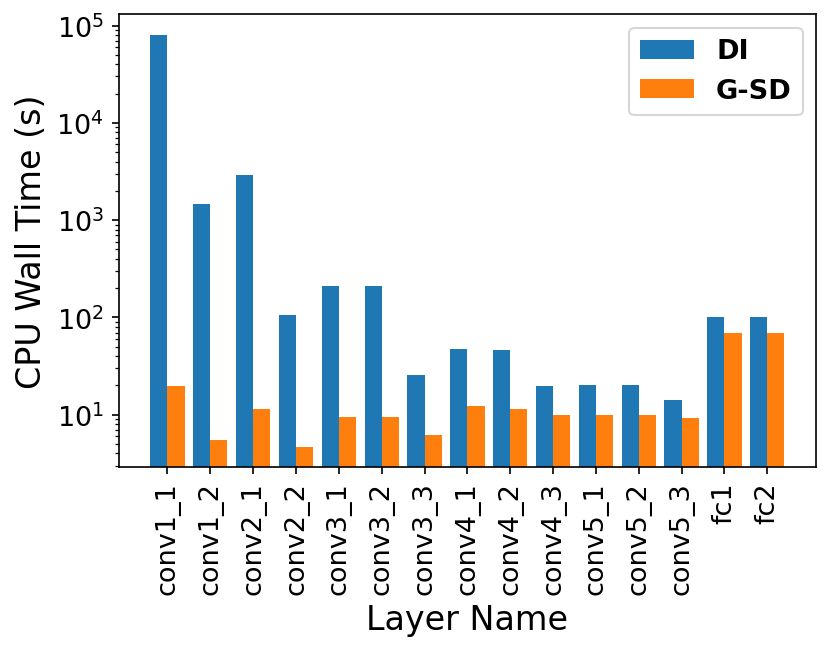}
    
    \caption{
        The CPU wall time for scoring all channels in VGG-16 layers by \metric\ and DI on ILSVRC-2012.
    }
    
    \label{fig:time_complexity}
\end{figure}

We conduct an empirical study to investigate the time complexity of two discriminant functions, \metric\ and DI, for channel scoring.
We pick 3,000 ILSVRC-2012 samples and compute their feature maps in VGG-16.
The maps are computed via an NVIDIA Tesla P100 GPU 
and the scorings of feature maps are executed on an Intel Xeon E5-2680 v4 CPU.
We calculate the CPU wall time for scoring all the channels in each layer 
by the two discriminant functions (excluding the time to obtain the maps).
From Fig.~\ref{fig:time_complexity}, 
we note that \metric\ generally has much less time complexity than DI for each layer's channel scoring process.
Moreover, for layer $\mathrm{conv1\_1 }$ where the feature maps are of size 224$\times$224, \metric\ shows a 4000$\times$ speedup over DI.

\section{More Visualizations}\label{sec:visual}

We repeat the analysis in Section~\ref{sec:qualitative_channel_sel} with additional images from different classes (not shown in the main paper).
Visually speaking, we again observe that our \metric\ selects more class-discriminative channels compared to state-of-the-art metrics.

\begin{table}[H]
\begin{center}

\begin{tabular}{c c c c c}
\hline
Layers & Parameters & FLOPs & Parameters (Pruned \%) & FLOPs (Pruned \%)\\
\hline

& \multicolumn{2}{c}{Original VGG-16} 
& \multicolumn{2}{c}{\metric-B VGG-16} \\
\hline


conv1\_1 & 1.7K & 87M & 0.3K (82.8) & 14.9M (82.8)  \\
conv1\_2 & 36.9K & 1.85B & 3.9K (89.5) & 0.19B (89.5) \\
\hline

conv2\_1 & 73.7K & 0.92B & 21.1K (71.4) & 0.26B (71.4)  \\
conv2\_2 & 0.15M & 1.85B & 49.7K (66.3) & 0.62B (66.3)\\
\hline

conv3\_1 & 0.29M & 0.92B & 92.7K (68.6) & 0.29B (68.6) \\
conv3\_2 & 0.59M & 1.85B & 97.8K (83.4) & 0.31B (83.4)\\
conv3\_3 & 0.59M & 1.85B & 0.17M (71.7) & 0.52B (71.7) \\
\hline

conv4\_1 & 1.18M & 0.92B & 0.43M (63.7) & 0.34B (63.7)\\
conv4\_2 & 2.36M & 1.85B & 0.52M (77.9) & 0.41B (77.9) \\
conv4\_3 & 2.36M & 1.85B & 0.67M (71.6) & 0.53B (71.6) \\
\hline

conv5\_1 & 2.36M & 0.46B & 1.47M (37.5) & 0.29B (37.5) \\
conv5\_2 & 2.36M & 0.46B & 2.02M (14.5) & 0.40B (14.5) \\
conv5\_3 & 2.36M & 0.46B & 2.02M (14.4) & 0.40B (14.4) \\
\hline

fc\_1 & 102.8M & 0.10B & 102.8M (0) & 0.10B (0) \\
fc\_2 & 16.8M & 16.8M & 16.8M (0) & 16.8M (0) \\
fc\_3 & 4.1M & 4.1M & 4.1M (0) & 4.1M (0) \\
\hline

Total & 138.34M & 15.47B & 131.19M (5.2) & 4.68B (69.7) \\
\hline

\end{tabular}

\end{center}
\caption{Detailed structure of the reported VGG-16 pruned model, \metric-B (Top1-Accuracy: 71.26\%). 
Fully connected layers are not pruned 
as 99.3\% of the FLOPs are in the convolutional layers, 
which makes FLOPs pruning of fully connected layers less cost-effective.}
\label{tab:VGG-16-Model}
\end{table}

\begin{table}[H]
\begin{center}

\begin{tabular}{c c c c c}

\hline
Blocks & Parameters & FLOPs & Parameters (Pruned \%) & FLOPs (Pruned \%) \\
\hline

& \multicolumn{2}{c}{Original ResNet-50} 
& \multicolumn{2}{c}{\metric-B ResNet-50} \\
\hline

Conv1 & 9.4K & 0.12B & 9.4K (0) & 0.12B (0)  \\
\hline

ResBlock1\_1 & 73.7K & 0.23B & 42.5K (42.3) & 0.13B (42.3)  \\
ResBlock1\_2 & 69.6K & 0.22B & 10.9K (84.3) & 34.3M (84.3) \\
ResBlock1\_3 & 69.6K & 0.22B & 25.5K (63.3) & 80.1M (63.3) \\
\hline

ResBlock2\_1 & 0.38M & 0.37B & 0.30M (20.8) & 0.28B (25.2) \\
ResBlock2\_2 & 0.28M & 0.22B & 49.9K (82.1) & 39.1M (82.1) \\
ResBlock2\_3 & 0.28M & 0.22B & 63.8K (77.1) & 50.0M (77.1) \\
ResBlock2\_4 & 0.28M & 0.22B & 93.0K (66.6) & 72.9M (66.6) \\
\hline

ResBlock3\_1 & 1.51M & 0.37B & 1.38M (8.7) & 0.33B (10.6) \\
ResBlock3\_2 & 1.11M & 0.22B & 0.38M (65.7) & 74.9M (65.7) \\
ResBlock3\_3 & 1.11M & 0.22B & 0.31M (72.2) & 60.8M (72.2) \\
ResBlock3\_4 & 1.11M & 0.22B & 0.30M (73.2) & 58.6M (73.2) \\
ResBlock3\_5 & 1.11M & 0.22B & 0.43M (61.4) & 84.2M (61.4) \\
ResBlock3\_6 & 1.11M & 0.22B & 0.69M (38.0) & 0.14B (38.0) \\
\hline

ResBlock4\_1 & 6.03M & 0.37B & 5.78M (4.2) & 0.36B (4.3) \\
ResBlock4\_2 & 4.46M & 0.22B & 3.72M (16.5) & 0.18B (16.5)\\
ResBlock4\_3 & 4.46M & 0.22B & 3.96M (11.1) & 0.19B (11.1)\\
\hline

fc1 & 2.05M & 2.05M & 2.05M (0) & 2.05M (0) \\
\hline

Total & 25.50M & 4.09B & 19.59M (23.2) & 2.28B (44.3) \\
\hline

\end{tabular}

\end{center}
\caption{Detailed structure of the reported ResNet-50 pruned model, \metric-B (Top1-Accuracy: 75.85\%).
We choose not to prune 
$\mathrm{Conv1}$ and $\mathrm{fc1}$ 
for the ease of implementation.}
\label{tab:ResNet-50-Model}
\end{table}

\begin{figure}[H]
    \centering
    \includegraphics[width=0.9\textwidth]{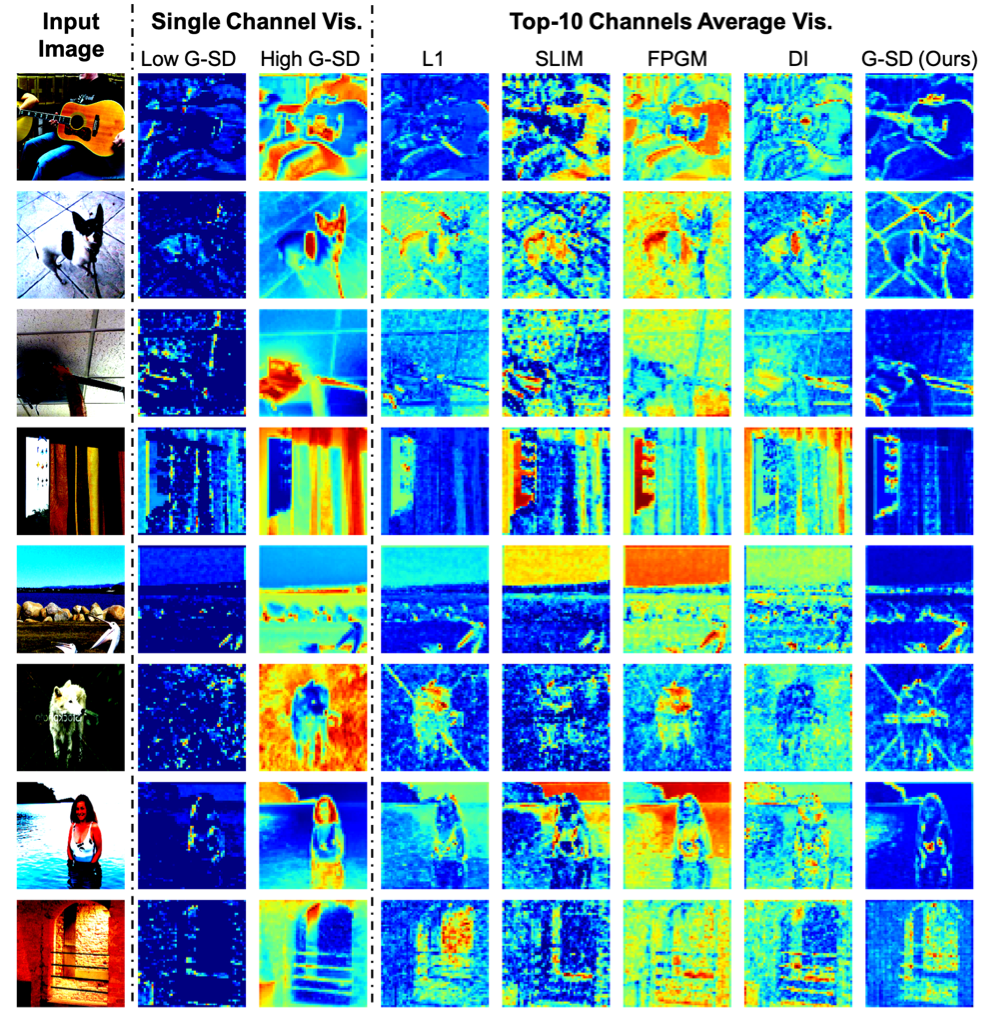}
    
    \caption{
       More visualizations of the qualitative channel selection analysis. 
    {\bf Col. 1:} Input images. 
    {\bf Col. 2-3}: Single channel visualization on channels with low and high \metric\ values. 
    {\bf Col. 4-8}: Average responses of the top-10 channels selected by different metrics. From left to right, the metrics are: $\ell$1-norm~\cite{li2016pruning}, batch-norm scaling factor~\cite{liu2017learning}, filter's geometric median~\cite{he2019filter}, DI~\cite{kung2019methodical}, and \metric.
    }
    
    \label{fig:Metric_Comparison}
\end{figure}

\end{document}